\documentclass[conference]{IEEEtran}
\usepackage{times}

\usepackage[numbers]{natbib}
\usepackage{multicol}
\usepackage{amssymb}
\usepackage{amsmath}
\usepackage[bookmarks=true]{hyperref}
\usepackage{graphicx}
\usepackage{subcaption}

\begin{document}

\title{Learning Models for Shared Control of Human-Machine Systems with Unknown Dynamics}

\author{\authorblockN{Alexander Broad\authorrefmark{1}\authorrefmark{3},
Todd Murphey\authorrefmark{2} and
Brenna Argall\authorrefmark{1}\authorrefmark{2}\authorrefmark{3}}
\authorblockA{\authorrefmark{1}Department of Electrical Engineering and Computer Science\\
\authorrefmark{2}Department of Mechanical Engineering \\
Northwestern University,
Evanston, Illinois 60208\\ Email: alex.broad@u.northwestern.edu}
\authorblockA{\authorrefmark{3}Rehabilitation Institute of Chicago, Chicago, Illinois 60611}}

\maketitle

\begin{abstract}
We present a novel approach to \textit{shared control} of human-machine systems.  Our method assumes no \textit{a priori} knowledge of the system dynamics.  Instead, we learn both the dynamics and information about the user's interaction from observation through the use of the Koopman operator.  Using the learned model, we define an optimization problem to compute the optimal policy for a given task, and compare the user input to the optimal input.  We demonstrate the efficacy of our approach with a user study.  We also analyze the individual nature of the learned models by comparing the effectiveness of our approach when the demonstration data comes from a user's own interactions, from the interactions of a group of users and from a domain expert.  Positive results include statistically significant improvements on task metrics when comparing a user-only control paradigm with our shared control paradigm.  Surprising results include findings that suggest that individualizing the model based on a user's own data does not effect the ability to learn a useful dynamic system.  We explore this tension as it relates to developing human-in-the-loop systems further in the discussion.
\end{abstract}

\IEEEpeerreviewmaketitle

\section{Introduction}
\label{sec-intro}

Complex human-machine systems are an increasing presence in many aspects of daily life.  As we continue to adopt these systems into our lives, it will become important to develop methods that share control between a robot and human operator. At the lowest level of control, the foremost question we must answer is how much should a person be allowed to interact with a controlled machine, particularly if that system is safety critical.  As we increase restrictions on the user's interaction, we can also hope to increase our certainty of the system's stability and safety; however, we also presumably decrease the positive effects of incorporating the human into the system in the first place.  For example, when piloting aircraft the human operator remains the primary (and often only) analyzer of dynamic environments, tasking the autonomous system with lower-level control tasks, like achieving desired way-points along a trajectory.  If we remove the pilot's ability to provide input to the aircraft, we may be able to ensure general stability of the system, but we may also reduce its overall capacity to respond in unexpected situations.  

It is our task, then, to develop techniques that can \textit{dynamically} adjust and allocate control authority to the human and robot partner in a manner that improves the safety and stability of the overall system.  This goal can be particularly challenging as the decision of how much control authority to allocate depends on the individual capabilities of each user.

Therefore, our goal is to develop a control sharing methodology that not only accounts for the system dynamics, but also how a user interacts with that system.  Importantly, we would like our approach to be valid for any mechanical device, and therefore we do not make use of any explicit model or \textit{a priori} knowledge of the dynamics.  Instead, we solve this problem with a data-centric technique by learning a joint model of the system dynamics and the user interaction, as represented by the Koopman operator \cite{koopman1931hamiltonian}.  The Koopman operator allows us to learn a (nominally infinite dimensional) linear model of any dynamic system through a spectral analysis of observations collected during system use. 

We then define our shared control methodology by integrating the learned model into a framework that uses Model Predictive Control (MPC) to provide outer-loop stabilization on the joint human-robot system.  Prior work \cite{fitzsimons2016optimal, tang2009stability} has demonstrated the efficacy of using optimal control as an outer-loop stabilization technique.  The unique aspect of our work is that the outer-loop stabilization is applied to a model learned directly from user data.  In essence, the model learned using the Koopman operator represents a computational understanding of both the human and mechanical systems.  We then use methods from optimal control to augment the user input and improve stability of the overall system.  This framework is sufficiently general that it could be applied to any robotic system with a human in the loop where the combined system can be expected to be differentiable.  

We provide related work and background on the Koopman operator in Section~\ref{sec-background}.  We then detail the scope of our problem and explain our approach in Section~\ref{sec-approach}.  Our experimental system is outlined in Section~\ref{sec-exper}.  The results are presented in Section~\ref{sec-results}, which we analyze and discuss in Section~\ref{sec-discuss}.  Finally, we conclude in Section~\ref{sec-conclusion}.

\section{Background and Related Work}
\label{sec-background}

This section presents related work in the shared control literature for human-machine systems.  We also provide a brief background on the Koopman operator and discuss some related work on its use in learning system dynamics.

\subsection{Shared Control}

Our motivation stems from a desire for a deeper understanding of how human-robot teams interact.  In particular, we are interested in developing a methodology that allows us to dynamically adjust the amount of control authority given to the robot and human partners \cite{hoeniger1998dynamically, hoffman2004collaboration}.  If done intelligently, and with appropriate knowledge of the individual capabilities of each team member, we can improve the overall efficiency, stability and safety of the joint system.  Approaches to shared control range from pre-defined, discretely adjustable methods \cite{kortenkamp2000adjustable} to probabilistic models \cite{javdani2015shared} to policy blending \cite{dragan2013policy}.  In addition to the original control signal space, shared control has been researched through haptic control \cite{nudehi2005shared} and compliant control \cite{kim1992force}, and studied in numerous domains.

In assistive and rehabilitation robotics, researchers have explored the effects of shared control on teleoperation of smart wheelchairs \cite{erdogan2017effect, trieu2008shared} and robotic manipulators \cite{kim2006continuous}.  Similarly, researchers have explored shared control as it applies to the teleoperation of larger mobile robots and human-machine systems, such as cars \cite{dewinter11smc} and aircraft \cite{matni08acc}.  When dealing with systems of this size, safety is often a primary concern.  

Mirroring our own experience with teams comprised of human partners, researchers have also explored the use of \textit{human trust} as a metric for improving shared control in human-robot teams \cite{desai12thesis}.  For example, researchers have developed systems that estimate the degree to which the human and robot are consistent in their actions and then updates the behavior of the robot to better align with the human \cite{xu2012icra}. Other's have described a bi-direction \textit{human-robot trust} metric based on task performance \cite{saeidi2016trust} and even developed a task-agnostic notion of \textit{robot trust} that is instead based on the user's understanding of the dynamic system \cite{broad2016trust}.  

The above works are conceptually similar to our own as they explore the question of how automation can be used to adjust to, and account for, the specific capabilities of the human partner.  However, in this work, we do not augment the user's control based on an explicitly computed metric.  Instead, we use observations of the user demonstrations to build a model of the joint human-robot system, and in this way, the effect of the individual user on the shared control system is implicitly encoded in the model learned from their interactions.

\subsection{The Koopman Operator}

In this work, we opt to learn the both the \textit{system dynamics} and information about the \textit{user interaction} directly from data.  This is a particularly important aspect of our approach as the resulting shared control methodology is \textit{generalizable to any dynamic system}, and \textit{specific to each user}.  Therefore, our shared control system is not based on a set of pre-defined, task-specific metrics or goals. Instead we simply capture relevant user-specific data from observations.  To achieve this, we use the Koopman operator \cite{koopman1931hamiltonian}.  We now provide the reader with a short technical introduction to this concept.

The Koopman operator is an infinite-dimensional linear operator that can capture all relevant information about any nonlinear dynamical system \cite{koopman1931hamiltonian}.  The mathematical theory states that we can describe the dynamics of any \textit{non-linear system} using a \textit{linear transformation} in a Hilbert space representation of the system state.  To rephrase this statement, the Koopman operator allows us to learn a linear mapping between a system's current state $x_t$ and the following state $x_{t+1}$, even when the relationship is non-linear in the initial state space.  Since the Koopman operator's definition, researchers have developed numerous algorithms capable of learning an approximation to the Koopman operator from observations of a dynamic system.  In this work, we use the Extended Dynamic Mode Decomposition (EDMD) algorithm, as described by \citet{williams2015data}, to approximate the Koopman operator and therefore use notation consistent with their work.

To define the Koopman operator, we consider a dynamic system $(\mathcal{X}, t, F)$ where $\mathcal{X} \subseteq \mathbb{R}^N$ is the state space, $t \in \mathbb{R}$ is time and $F : \mathcal{X} \rightarrow \mathcal{X}$ is the state evolution operator.  The Koopman operator, $\mathcal{K}$, then is defined as the composition of $\phi$ with $F$, such that 

\begin{equation}
\mathcal{K} \phi = \phi \circ F
\label{koopman-eq}
\end{equation}

\noindent where $\phi$ represents an infinite dimensional basis function.

The Hilbert space transformation is defined by a basis function, $\phi : \mathcal{X} \rightarrow \mathbb{C}$, which transforms the standard state definition into a Hilbert space.  In practice, this basis function is often a non-linear combination of the initial state representation and is used to approximate the Koopman operator.  This basis function can be significantly larger in dimensionality than the initial state space.  Common choices for the basis function include Hermite polynomials and radial basis functions \cite{williams2015data}.  By acting on the Hilbert state representation, the \textit{linear} Koopman operator is able to capture the complex, nonlinear dynamics described by the evolution operator.

While the theory behind the Koopman operator \cite{koopman1931hamiltonian} was developed in the early 1930's, it has mostly been ignored as a method of solving dynamic equations in favor of geometric methods.  Recently, however, there has been renewed interest in the approach as it fits naturally into the modern \textit{big data} world \cite{budivsic2012applied}.  In addition to learning a model from observations, the Koopman theory extends easily to higher-dimensional systems, often a difficult aspect of standard geometric methods.

In contemporary work, the Koopman operator has been successfully used to learn the dynamics of numerous challenging dynamic systems as wide ranging as fluid flow \cite{schmid2010dynamic}, the stock market \cite{hua2016using} and the brain \cite{brunton2016extracting}. 

\section{Problem Definition and Approach}
\label{sec-approach}

\begin{figure}
\centering
\includegraphics[width=\hsize]{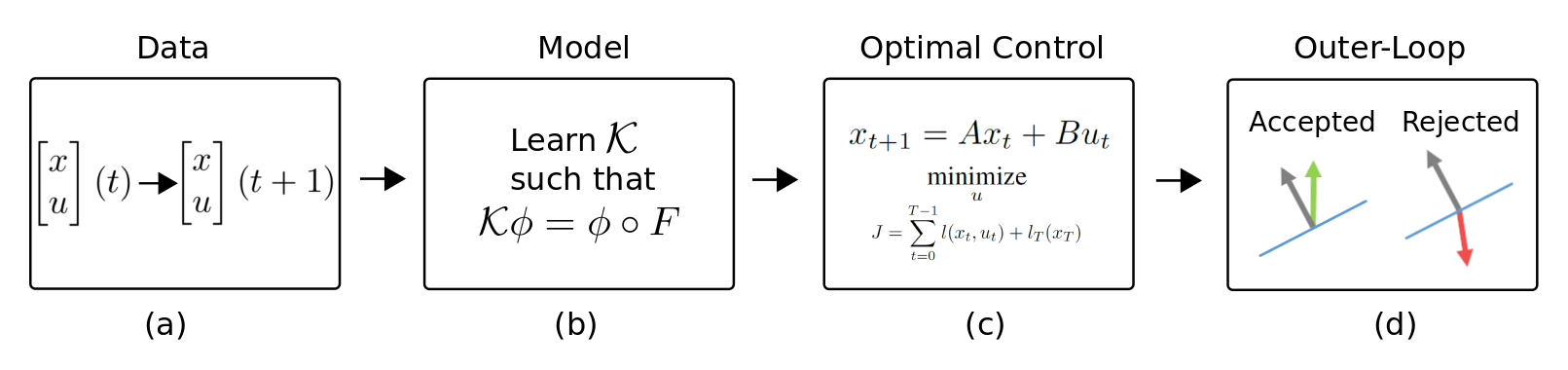}
\caption{Pictorial description of the proposed workflow.  (a) Capture data from user interaction with the dynamic system.  (b)  Learn model from observations using Koopman operator.  (c)  Compute optimal control response to current state using the learned model.  (d) Define shared control system that compares optimal control response (gray) and user input (green/red).}
\label{fig:approach}
\end{figure}

In this work, our primary goal is to develop a shared control methodology that improves the stability of any complex dynamic system without relying on a set of hand-picked features to describe the relationship between the human and the automation, and without any \textit{a priori} knowledge of the system or user.  Toward that end, we take inspiration from learning-based approaches and develop a model of the joint human-robot system based solely on observations of the user.  More specifically, we allow the user to interact with and control the system during a data collection phase.  We then use the EDMD algorithm to approximate the Koopman operator describing the joint system dynamics. To close the loop on our shared control approach, we incorporate Model Predictive Control as an outer-loop stabilization technique which we apply to the joint human-robot system.  While our approach separates the data collection and model learning phases, recent work has demonstrated the efficacy of using the Koopman operator in an online learning paradigm \cite{delatorre2017model}.

As described, the model used to compute the optimal control solution is learned directly from user interaction data via the EDMD algorithm.  Importantly, this approach \textit{does not require} a prior model of the system dynamics (which can be challenging to define for complex systems), information about the user's ability (which we instead learn from observation, along with the dynamics) or hand-picked features that relate to the human-robot interaction (which otherwise may be task-specific and generalize poorly to novel environments).  Other learning-based MPC controllers have also been proposed \cite{aswani2013provably, lenz2015deepmpc}, however they do not incorporate the influence of the human in the learning process or in the final control.  Our approach to shared control is outlined in Figure~\ref{fig:approach} and described in detail in the following subsections.

\subsection{Compute Koopman Operator Based on User Observations}

The first step in our approach requires the collection of data.  Specifically, by collecting data of a given user's interaction with the mechanical system, we can compute the Koopman operator that describes the dynamics of the joint human-robot system.   Similar to \citet{proctor2016generalizing}, we incorporate the user input directly into the state so that we have $x_{t+1} = f(x_t, u_t)$.  We then define a basis function and compute the solution to the linear least squares problem in the Hilbert space as described in \cite{williams2015data}.  In this work we do not compute the full Koopman operator and instead \textit{assume} linear dynamics of the combined system, which yields a simplified linear structure, allowing us to only compute a finite spectrum.  For a system that is expected to be nonlinear, the only necessary change is to incorporate more basis functions. 

\subsection{Outer-Loop Stabilization via Optimal Control and the Koopman Operator}
\label{sec:mpc_with_koop}

The second step in our approach is to define a shared control framework that uses optimal control as an outer-loop stabilization technique \cite{fitzsimons2016optimal}.  By requiring the user input to remain within a small deviation of the optimal solution computed by the MPC algorithm, we ensure that the resulting control will not destabilize the system and that the user will remain otherwise unobstructed by their autonomous partner.

To determine the optimal control input, $u$, we define the Model Predictive Control problem as

\begin{equation}
x_{t+1} = Ax_t + Bu_t
\label{eqn-mpc}
\end{equation}

\noindent where $A \in R^{mxm}$ describes the system dynamics, $B \in R^{mxp}$ describes the effect of the user input, $m$ is the dimension of the state space and $p$ is the dimension of the control input.  Then, to compute the optimal control sequence, we must solve the following optimization

\begin{equation*}
\begin{aligned}
& \underset{u}{\text{minimize}}
& & J = \sum_{t=0}^{T-1} l(x_t,u_t) + l_T(x_T) \\
& \text{subject to}
& & x_{t+1} = Ax_t + Bu_t 
\end{aligned}
\end{equation*}

\noindent where $l$ and $l_T$ are the running and terminal cost, respectively.

Importantly, in this work we have assumed no a priori knowledge or model of the system dynamics.  For that reason, we must incorporate the model we learned via EDMD into the MPC framework.  To transform the problem into a form that incorporates the Koopman, we take our model, defined as 

\begin{equation}
x_{t+1} = \mathcal{K}^T\phi(x_t,u_k) 
\label{eqn-koopman}
\end{equation}

\noindent and re-write it using the form we defined in Equation~\ref{eqn-mpc}.  To do so, we linearize Equation~\ref{eqn-koopman} such that

\begin{equation*}
x_{t+1} = \mathcal{K}^T \frac{\partial \phi}{\partial x} x_t +  \mathcal{K}^T \frac{\partial \phi}{\partial u} u_t\\
\end{equation*}

\noindent which gives us the desired form where 

\begin{equation}
A = \mathcal{K}^T \frac{\partial \phi}{\partial x},~B = \mathcal{K}^T \frac{\partial \phi}{\partial u}.
\end{equation}

By solving the MPC problem defined using the learned Koopman operator model, we compute the optimal control input at a given state.  Optimal control has previously been demonstrated as a viable outer-loop control mechanism \cite{fitzsimons2016optimal}.  

To close the loop, we compare the user input to the optimal input at each time-step.  If the signal is close enough to the optimal control input, we allow the signal through to the system, otherwise we provide no input to the system.  Depending on the specific criteria we use to decide which user signals are let through to the system, we can provide guarantees on the stability and likelihood of task success.  In this work we restrict the user input to the system to be in the same half-plane as the optimal control solution and place no other limitations on the human-machine interaction.  

\section{Experimental Validation}
\label{sec-exper}

The proposed shared-control framework is validated on a simulated lunar lander (see Figure~\ref{fig:ll_env}).  The dynamic system is a modified version of the open-source environment implemented in the Box2D physics engine and released by OpenAI \cite{brockman2016gym}.  Our modifications (1) allow for multi-input user control via the keyboard or a joystick, and (2) incorporate the codebase into the open-source ROS framework.

The experiments were run on a Core i7 laptop with 8 GB of RAM.  In each trial, the lunar lander was initialized to the same $x, y$ position, to which we added a small amount of Gaussian noise ($\mu = 0.2$).  Additionally, a random acceleration was applied at the start of each trial.  The goal location was constant throughout all trials.  The operator used a PS3 controller to interact with the system.  The joystick controlled by the participant's dominant hand fired the main thruster, and the opposing joystick fired the side thrusters.  As the user moved through the environment, we kept track of the full state space at each timestep.  The study consisted of 16 total participants (11 female, 5 male).

\subsection{Environment and State Space}

\begin{figure}[t]
\centering
\fbox{\includegraphics[width=0.75\hsize]{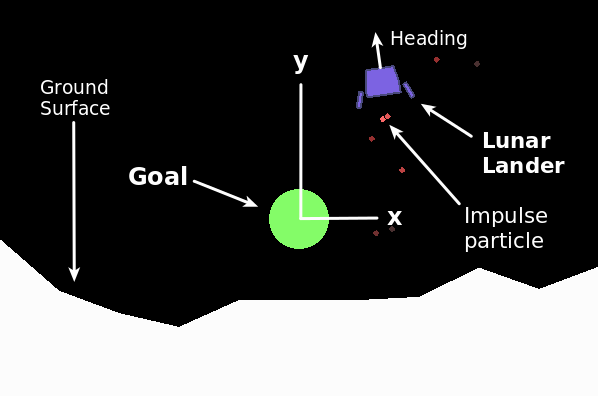}}
\caption{Simulated lunar lander system.  The green circle is the visualization of the desired goal location.  The red dots represent an engine firing.}
\label{fig:ll_env}
\end{figure}

We chose to use a simulated lunar lander (rocket) system as our experimental environment for a number of reasons.  First, this environment is extremely challenging for a novice user, but can be improved upon (and sometimes mastered) given enough time and experience.  Similar to a real rocket, one of the main control challenges is the stability of the system.  As the rocket rotates along its yaw axis, firing the main thruster can produce nonintuitive dynamics for a novice.  Furthermore, once the rocket has begun to rotate, momentum can easily overwhelm a novice user who is unfamiliar with such systems.  Therefore, it is often imperative, particularly for non-expert users, to maintain a high degree of stability at all times in order to successfully achieve the task.  The goal in this environment requires the user to navigate the lander from its initial location to the goal location (represented by the green dot in Figure~\ref{fig:ll_env}).  To complete the experiment, the lunar lander's heading must be nearly perpendicular to the ground plane and the linear and rotational velocities near zero.  In addition to the control challenges, we chose this environment because the simulator abstracts the system dynamics through calls to the Box2D physics engine; therefore, we do not have an exact model and thus have an explicit need to learn one.

The lunar lander is defined by a 6D state space ($\mathcal{X}_{ll}$) made up of the position ($x,y$), heading ($\theta$), and their rates of change ($\dot{x}, \dot{y}, \dot{\theta}$).  Therefore, we have
\begin{equation*}
\mathcal{X}_{ll} = [x, y, \theta, \dot{x}, \dot{y}, \dot{\theta}].
\end{equation*}

The control input to the system is a continuous two dimensional vector which represents the throttle of the main and rotational thrusters.  The main engine can only apply positive force.  The left engine fires when the second input is negative, while the right engine fires when the second input is positive.  The main engine applies an impulse that acts on the center of mass of the lunar lander, while the left and right engines apply impulses that act on either side of the rocket's body.  

We remind the reader that our goal is to learn both the system dynamics and user interaction.  For this reason, we must collect data both on the system state and also the control input.  Then, to compute the Koopman operator for a given user, we concatenate the state space of the lunar lander with the user input.  Together, this defines an eight dimensional system
\begin{equation*}
\mathcal{X} = [x, y, \theta, \dot{x}, \dot{y}, \dot{\theta}, u_{main}, u_{rot}]
\end{equation*}

\noindent where the first six terms define the lunar lander state and $u_{main}, u_{rot}$ are the main and rotational thruster values, through which the user interacts with the system.

For the described system we choose a simple basis function, $\phi$, comprised of a bias term, the system states and the control states.  We then use the Koopman spectral method to compute a model of the joint human-robot system.  As we restrict the focus of our model to the linear states, we can solve the resulting MPC problem by using a Linear Quadratic Regulator (LQR) in the original state space.  We can recover the original state definition by pulling the values out of the updated Hilbert space representation.  We can then compute the optimal control input by using the  discrete algebraic Riccati equation to solve the LQR.  Finally, we compare the optimal control input to the user control input in each dimension, and remove user inputs that are in the opposite half-plane to the optimal control input.

\begin{figure*}
    \centering
    \begin{subfigure}[t]{0.245\hsize}
        \centering
        \includegraphics[width=\hsize]{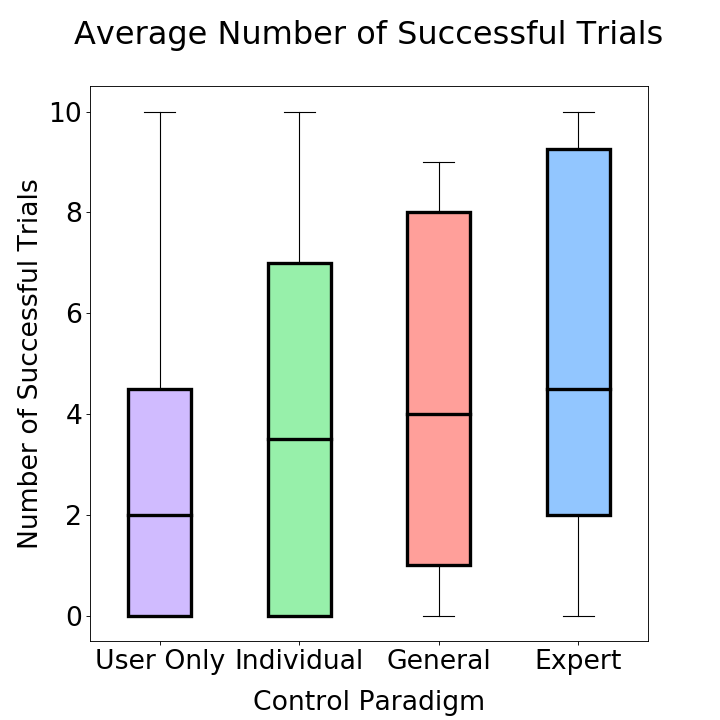} 
        \caption{}
        \label{fig:res-a}
    \end{subfigure}
    \begin{subfigure}[t]{0.245\hsize}
        \centering
        \includegraphics[width=\hsize]{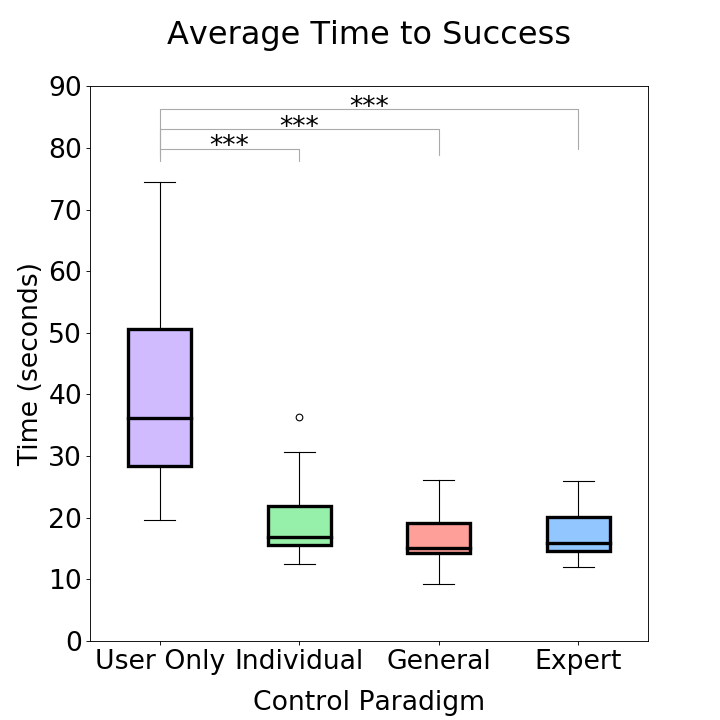}
        \caption{} 
        \label{fig:res-b}
    \end{subfigure}
    \hfill
    \begin{subfigure}[t]{0.245\hsize}
        \centering
        \includegraphics[width=\hsize]{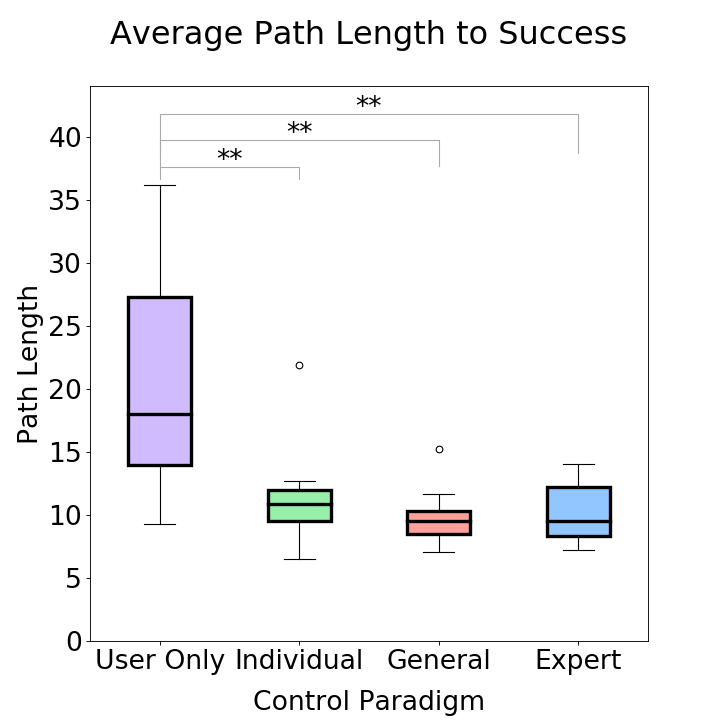} 
        \caption{}
        \label{fig:res-c}
    \end{subfigure}
    \begin{subfigure}[t]{0.245\hsize}
    	\centering
        \includegraphics[width=\hsize]{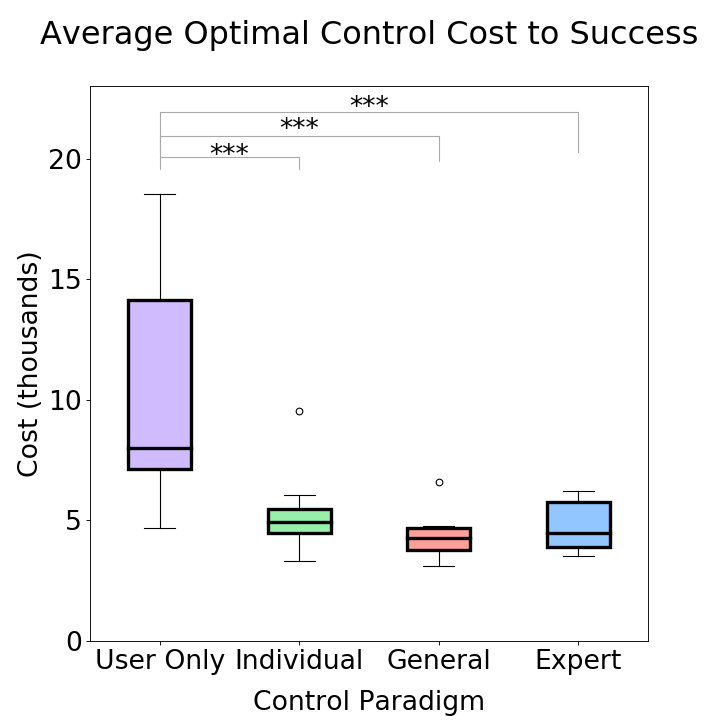} 
        \caption{}
        \label{fig:res-d}
    \end{subfigure}
    \caption{(\subref{fig:res-a})~Number of successful trials under each control paradigm.  Despite a visual trend, we find no statistically significant difference between the user-only control paradigm and the shared control paradigms. (\subref{fig:res-b})~Average time to successfully complete a trial under each control paradigm. (\subref{fig:res-c})~Average path length taken to successfully complete a trial under each control paradigm.  (\subref{fig:res-d}) Average total optimal control cost computed during a successful trial under each paradigm.  For all three metrics (b-d), post-hoc pair-wise t-tests using Holm-Bonferonni corrected alpha values find statistically significant differences between the user-only paradigm and each shared control paradigm (time: $p < 0.005$, path length: $p < 0.01$, cost: $p < 0.005$).   Additionally, our post-hoc t-tests find no statistically significant difference between any of the shared control paradigms.}
    \label{fig:avg-secondary-metrics}
\end{figure*}

\subsection{Experimental Protocol}

The goal of each experimental trial is to maneuver the lunar lander from its initial state to the goal state.  The $x,y$ position of the goal state is displayed to the user with a green circle.  A trial is over either when the center of an upright lunar lander is fully contained within the green circle and the linear and angular velocities are near zero, or when the lander moves outside the bounded environment or crashes into the ground.

To study the effect of our shared control system, we compare four distinct paradigms.  In the first paradigm, the user is in full control of the lander and is not assisted by the autonomy in any way; we call this approach \textit{User Control}.  In the remaining three paradigms the autonomy provides outer-loop stabilization on the user's input as described in Section~\ref{sec:mpc_with_koop}.  The main distinction between the remaining approaches is the model that is used to compute the outer-loop control.  Specifically, in the second paradigm the model is defined by a Koopman operator learned on data captured from earlier observations of the current user; we call this approach \textit{Individual Koopman}.  In the third paradigm, the model is defined by a Koopman operator learned on data captured from observations of three novice participants prior to the experiment (who were not included in our analysis); we call this approach \textit{General Koopman}.  In the fourth paradigm, the model is defined by a Koopman operator learned on data captured from observations of an expert user (the first author of the paper, who had significant practice controlling the simulated system); we call this approach \textit{Expert Koopman}.

The four control paradigms are chosen to (1) evaluate the efficacy of our joint control system as opposed to a natural learning scenario without any automation, and (2) to evaluate the individual nature of the learned system dynamics.  We analyze the effect of our shared control system by comparing the \textit{User Control} paradigm to each of the shared control paradigms.  We analyze the individual nature of the learned models by comparing the results under the \textit{Individual Koopman}, \textit{General Koopman} and \textit{Expert Koopman} paradigms.

Each experiment begins with a training period for the user to become accustomed to the dynamics of the system and the interface. This training period continues either until the user is able to successfully achieve the task three times in a row, or 15 minutes elapse.  During the next phase of the experiment, we observe the user and collect data from 10 trials, which we then use to compute the Koopman operator.  Finally, each user performs the task under the four control paradigms detailed above (10 trials each).  The order in which the control paradigms are presented to the user is counter-balanced so as not to bias the results.   

\section{Results} 
\label{sec-results}

Our analysis investigates the efficacy of the proposed framework for developing a platform-agnostic, user-specific shared control methodology.  Specifically, we evaluate our ability to learn a model of the joint human-robot system using the Koopman operator, and subsequently, the effectiveness of MPC as an outer-loop stabilization technique when applied to the learned dynamic model.  We also analyze the individual nature of the Koopman representation of the system dynamics by comparing models learned from a user's own data to a model learned from a committee of novices and a model learned from a domain expert.

\subsection{Performance Differences Between Control Paradigms}

We begin by comparing how well our methodology is able to assist in stabilizing the system dynamics as compared to a standard learning approach where the user gets no assistance and instead improves with experience.  To do so, we evaluate the ability of each user to successfully complete the desired task under each shared control paradigm.  Our results are based on statistical tests comparing the performance of participants under the user-only control paradigm to participants under each shared control paradigm, along a set of pertinent metrics.  Our experiment consists of 10 trials per user in each paradigm for a total of 160 trials per paradigm.  Our analysis consists of one-way ANOVA tests conducted to compare the effect of the shared control paradigms on each of the dependent variables.  These tests allow us to statistically analyze the effect of each paradigm while controlling for overinflated type I errors that are common with repeated t-tests.  Each test is computed at a significance value of 0.05.  When the omnibus F-test produces significant results, we conduct post-hoc pair-wise Student's t-tests using Holm-Bonferroni adjusted alpha values \cite{wright1992adjusted}.  We note that this type of analysis is used for \textit{all reported results.}  The post-hoc tests allow us to further evaluate the cause of the significance demonstrated by the ANOVA by comparing each pair of shared control paradigms separately.  Similar to the ANOVA test, the Holm-Bonferroni correction is used to reduce the likelihood of type I errors in the post-hoc tests.

The first metric we compute is the the success rate observed during our experiment.  We can interpret the success rate of a user, or shared control system, on a set of trials as a measure of skill.  The greater the skill, the higher the success rate.  The average number of successful trials are displayed in Figure~\ref{fig:res-a}.  While we do observe a positive trend in favor of the shared control paradigms, an analysis of variance showed that the effect of the shared control paradigm on the success rate was not significant (F(3, 59) = 1.32, p = 0.28).

The fact that we do not see a statistically significant difference along this metric is likely due to our chosen shared control methodology (Section~\ref{sec:mpc_with_koop}).  In particular, our approach does not supplement the user signal with information from the optimal control solution, instead the outer-loop control simply blocks signals that are too far from the optimal solution.  For this reason, the success of the joint system on a particular task is still highly dependent on the user's own understanding of the dynamic system and skill in controlling it.  Altering this parameterization, or simply replacing the choice of outer-loop control, will have many practical implications on the stability of the overall system and shared-control paradigm.

To further analyze any difference in the performance of the users over the four control paradigms, we compare a number of other relevant metrics.  Specifically, we analyze the average time, average path length and the average total running value of the optimal control cost function when successfully completing a trial.  These metrics are broken down by control paradigm and displayed in Figures~\ref{fig:res-b}, \ref{fig:res-c}, \ref{fig:res-d}.

Similar to the success rate, we can interpret the average time, path length, and the value of the running cost function as measures of skill.  That is, we believe skilled users, or shared control systems, will be able to achieve the task faster, produce more direct trajectories and spend less time in high cost areas than those with less skill.  An analysis of variance showed that the shared control paradigm had a significant effect on the trial time ($F(3, 44) = 15.39, p < 0.001$), the path length ($F(3, 44) = 12.26, p < 0.001$) and the MPC cost ($F(3, 44) = 15.40, p < 0.001$).  In each case, our post-hoc pair-wise Student t-tests using the Holm-Bonferroni correction found statistically significant differences between the performance of the users in the user-only control paradigm and users in the shared control paradigms based on the individual, general and expert datasets ($p < 0.005, p < 0.01, p < 0.005$, respectively).  We interpret these results as evidence that our shared control methodology improves the performance of the joint human-robot system.

\subsection{Personalization of Learned Models}

\begin{figure}[h]
\centering
\includegraphics[width=.55\hsize]{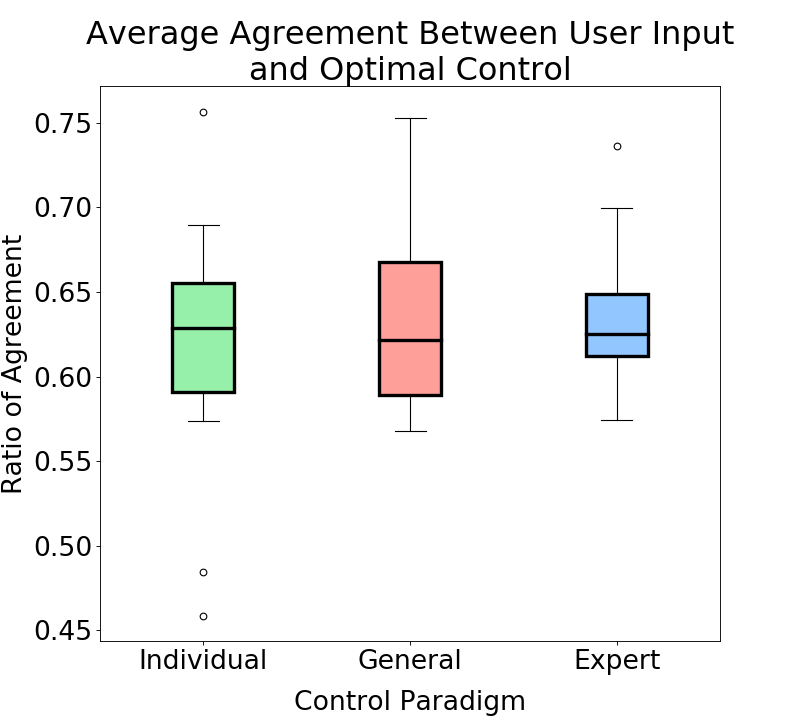}
\caption{Average agreement.}
\label{fig:avg-agreement}
\end{figure}

We also analyze the individual nature of the learned models by comparing the skill of each user under the \textit{Individual Koopman} with each user under the \textit{General Koopman} and \textit{Expert Koopman} paradigms.  We are particularly interested in whether there is a personalized interpretation of the participants' interactions with the outer-loop control.  To compare different models, we compute the average percentage of user inputs that agree with the optimal controller over the course of a trial.  The average agreement metric is broken down by control paradigm and presented in Figure~\ref{fig:avg-agreement}.  

An analysis of variance showed that the effect of the source of the model data on the average agreement was not significant ($F(2, 44) = 0.87, p = 0.43$).  Interestingly this suggests a uniformity in the response to system state across users in our study.  Initially, we hypothesized that the models may learn an individual understanding of the interaction between the system and user, however, our results seem to suggest that the system is adapting to the person, and not the other way around. 

In fact, a direct comparison of the linear dynamics, as represented by the learned A and B matrices demonstrated a \textit{striking similarity} between the learned representations across users.  We found an average standard deviation of 2.5$\%$ of the mean in the A matrices and an average standard deviation of 1.6$\%$ of the mean in the B matrices.  These extremely low values demonstrate that the users in our study act in essentially the same manner as each other when they interact with the system despite varying expertise. A caveat to this analysis is that it remains challenging to isolate and distinguish between which data source (i.e. the user input, system dynamics or physics) is responsible for the similarities and differences we observe between the learned models.

\begin{figure*}[t]
\centering
\includegraphics[width=\hsize]{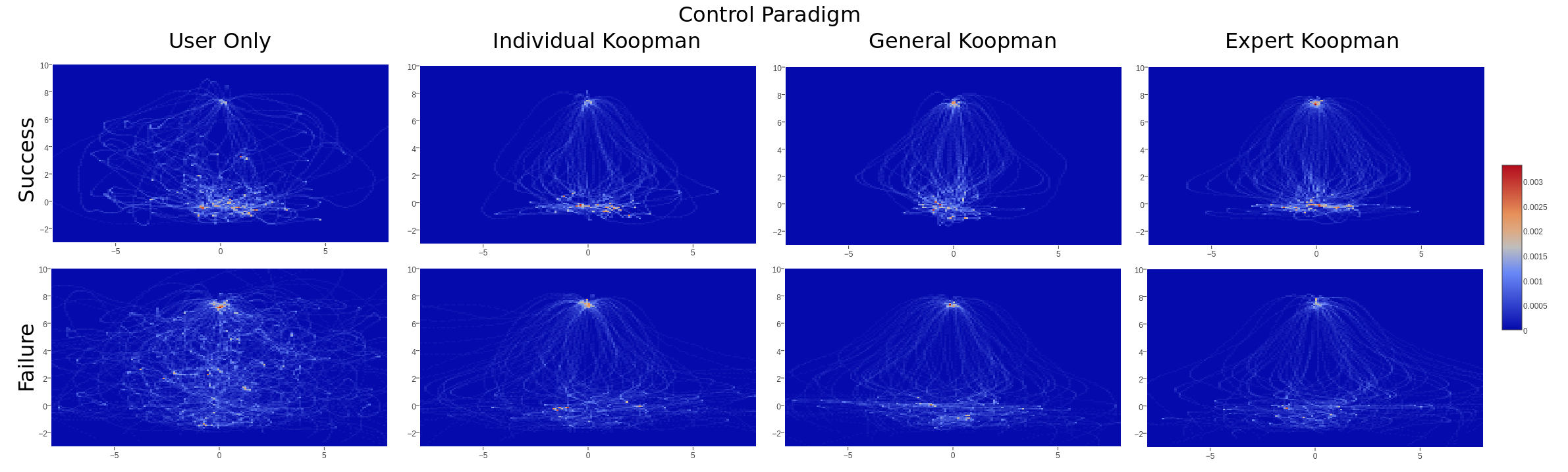}
\caption{Heat maps which visualize the most frequently visited parts of the state space.  The data is broken down by control paradigm (columns) and whether the trial was successful (rows).  The intensity (legend) represents the percentage of time spent in that location and is normalized across all trials.}
\label{fig:heatmaps}
\end{figure*}

One possible reason for this result (beyond the potentially obvious one, that there is no individualization in models learned using this method) is the highly linear nature of the dynamics of the chosen experimental system.  It is of note that our choice of a linear basis function excludes the non-linear terms between the user input and the state space, which may be useful in defining a more explicit relationship between the individual and the learned dynamics.  In future work, we plan to incorporate larger, non-linear basis functions.

There is, of course, a positive spin to this result: namely, that one can use models learned from an individual rather than having to maintain a large dataset of all potential users of the system.  As the task changes, this would be very helpful as there may not be a pre-formed library of responses available in the training data set.  Additionally, as we see no significant difference in task metrics between the different shared control paradigms, the results also suggest that the model can be learned from \textit{any} user instead of requiring demonstrations from a domain expert (something that is likely not true with other demonstration-based approaches \cite{argall2009survey}).  

\subsection{Density Analysis of Trajectories Generated Under Each Control Paradigm}

Finally, we also compare the different shared control methods through an analysis of the distribution of trajectories that we observe under each paradigm.  Figure~\ref{fig:heatmaps} depicts heat maps which represent the most frequently occupied sections of the state space.  The heat maps are separated so that we can analyze the data based on the control paradigm and whether the user was able to complete the task on a given trial.  We begin with a qualitative analysis of the heat maps.

The first distinction we draw is between the user-only control paradigm and \textit{all} of the shared control paradigms.  In particular, the distribution of trajectories in the user-only paradigm depict both larger excursions away from the target and lower levels of similarity between individual executions.  When we focus specifically on which parts of the state space users spend the most time in (as represented by the intensity of the heat map), we see two main clusters of high intensity (around the start and goal locations) in the shared control paradigm, whereas we see a wider spread of high-intensity values in the user-only control paradigm.  This suggests an increased level of control in the shared control paradigms.

The second distinction we draw focuses on a comparison between the successful and unsuccessful trials.  Specifically, we note that heat maps computed from the failed trials under the shared control paradigm demonstrate similar properties (e.g. the extent of the excursions away from the target, as well as two main clusters of intensity) to the heat maps computed from successful trials under the shared control paradigm.  This suggests that users may have been closer to succeeding in these tasks than the binary success metric gives them credit for.  By comparison, the heat map computed from the failed trials under the user-only control paradigm depicts a significantly different distribution of trajectories with less structure.  Specifically we observe numerous clusters of intensity that represent time spent far away from the start and goal locations.  This suggests that users were particularly struggling to control the system in these cases.

These observations are supported by an evaluation of the ergodicity \cite{mathew2011metrics, miller2016ergodic} of the distributions of trajectories described above.  To perform this comparison, we compute the ergodicity of each trajectory with respect to a probability distribution defined by a Gaussian centered at the goal location (which represents highly desirable states).  This metric can be calculated as the weighted Euclidean distance between the Fourier coefficients of the spatial distribution and the trajectory.

We first compare ergodicity between all paradigms, by analyzing \textit{all} the trajectories observed under each control paradigm.  To perform this analysis, we again compute a one-way ANOVA test.  An analysis of variance showed that the effect of the shared control paradigms on trajectory ergodicity is significant ($F(3, 635) = 12.46, p < 0.001$).  Our post-hoc tests find a statistically significant difference between the user-only control paradigm and each of the shared control paradigms ($p < 0.001$ for all cases).  We do not find the same statistical difference when we compare the shared control paradigms to each other.  This suggests that all the shared control paradigms are ergodic with respect to the goal location, whereas the user-only control paradigm is not.  We can interpret this result to mean that our shared control paradigms do indeed aid the user in successfully getting to the desired goal configuration, because users spent a significantly greater portion of their time near the goal location.

We further analyze the ergodicity results by separating the trajectories based on whether they come from a successful or unsuccessful trial.  By focusing on the subset of trajectories that come from unsuccessful trials, we can expand our examination of the efficacy of our shared control methodology beyond the task-specific metrics we compute during successful trials.  An analysis of variance shows that the effect of the shared control paradigm on the ergodicity of the trajectories was significant ($F(3, 368) = 6.59, p < 0.0005$).  Our post-hoc tests found that there was a statistically significant difference between the user-only control paradigm and each of the shared control paradigms ($p < 0.01$ for all cases).  Again, we do not find the same statistically significant difference between the ergodicity of the trajectories between any of the shared control paradigms.  These results suggest that the shared control paradigms are helpful in controlling the lunar lander even when the users are providing input that is ultimately unsuccessful in achieving the task.

We also analyze the subset of trajectories that come from successful trials, which allows us to focus on a comparison to the \textit{best} results demonstrated under the user-only control paradigm.  An analysis of variance shows that the effect of the shared control paradigm on the ergodicity of the trajectories was significant ($F(3, 262) = 8.33, p < 0.001$).  Again, our post-hoc tests find a statistically significant difference between the user-only control paradigm and the shared control paradigms based on the individual, general and expert datasets ($p = 0.02, p = 0.016, p = 0.02$, respectively).  This result suggests that the application of our shared control paradigm is helpful, even during the best of the user trials.

\section{Discussion} 
\label{sec-discuss}

The experimental results demonstrate that we are (1) successfully able to learn a model of the system dynamics and user interaction, and (2) can use the learned model in a shared control framework that significantly improves the performance of the joint human-robot system on a given task.  By developing a shared control framework that is learned from user observation, we are able to adapt the level of autonomous intervention to the skill of the user, thus creating a level of combined performance greater than the individual alone.

Additionally, by comparing the performance of a human operator when using a model learned on data they provide versus data provided by a group of other novices or an expert, we are able to evaluate the individual nature of the learned representation.  In contrast to our initial hypothesis, the direct interaction between the user and the outer-loop control (as measured by the average agreement) demonstrates a uniformity in the representation learned from different datasets.

There are a few choices that we made during the experiments that could be further evaluated.  For example, in this work we chose to use a linear basis function to approximate the Koopman operator.  In future work, we plan to expand our choice of basis function to higher-order, nonlinear basis functions, which may allow us to further evaluate the individual nature of the learned models.  Additionally, in this work we chose to learn the models from observations collected over the course of 10 trials.  However, anecdotally, we found that we needed significantly less data than that to learn the dynamics. How the amount of data used to compute the Koopman operator effects the efficacy of our approach is left for future work.  Finally we note that, in order to successfully learn the dynamics, the training data need not come from successful trials as defined by the task metric (in fact, this was rare in our experiments).  However, the training data also cannot come from random inputs.  Therefore, how the distribution of control input and state space exploration effects the efficacy of our methodology is also left for future work.

Additionally, when defining the conditions of the outer-loop stabilization, we chose simply to allow the user signal directly through to the system if their signal and the optimal control signal were in the same half-plane.  However, the magnitude of the two signals can still significantly differ and therefore the user can still provide sub-optimal control inputs.  In a similar vein, when the user signal and the optimal control disagree, we could choose to use the optimal control solution to stabilize the system.  We believe the downside to both of these choices is that it would be more challenging for the user to comprehend how they are directly influencing the system.  

\section{Conclusion} 
\label{sec-conclusion}

In this work, we investigate a novel shared control method that dynamically allocates control authority to a given user based on a learned model of the joint human-robot system.  One particularly important aspect of this work is that \textit{we do not rely on a priori knowledge, or a high-fidelity model, of the system dynamics}.  Instead, we learn the system dynamics \textit{and} information about the user interaction with the system directly from data.  We learn this model using the Koopman operator. 

Results from a user study demonstrate that incorporating the learned models into our shared control framework statistically improves the performance of the operator along a number of pertinent metrics.  Furthermore, an analysis of trajectory ergodicity demonstrated that our shared control framework was able encourage the human-machine system to spend a significantly greater percentage of time in desirable states.  In contrast with our initial hypothesis, we did not observe performance differences due to personalized models.  We leave further exploration of this idea to future work.

In conclusion, we believe that our approach is an effective step towards shared control of human-machine systems with unknown dynamics.  In particular, we expect that this approach is generalizable to any differentiable dynamic system.  As part of this work, we have \textit{open-sourced} the software used in these experiments.  This includes code that can be used to (1) learn the Koopman operator representation from collected data, (2) integrate the linearization of the model with our MPC framework, and (3) the experimental environment.  The code  can be found on the author's \href{https://github.com/asbroad/koopman_operator_model_learning}{github page}.

\section*{Acknowledgments}

This material is based upon work supported by the National Science Foundation under Grant CNS 1329891. Any opinions, findings and conclusions or recommendations expressed in this material are those of the authors and do not necessarily reflect the views of the aforementioned institutions.  We would like to thank Gerardo De La Torre and Ian Abraham for their many helpful discussions on the Koopman operator.  

\bibliographystyle{plainnat}
\bibliography{references}

\end{document}